\documentclass[runningheads]{llncs}
\usepackage[T1]{fontenc}
\usepackage{graphicx}
\begin{document}
\title{On Importance of Code-Mixed Embeddings for Hate Speech Identification}
%


\author{Shruti Jagdale\inst{1,3} \and
Omkar Khade\inst{1,3} \and
Gauri Takalikar\inst{1,3} \and
Mihir Inamdar\inst{1,3} \and
Raviraj Joshi\inst{2,3}}
\authorrunning{Shruti et al.}
%
\institute{Pune Institute of Computer Technology, Pune, India \and
Indian Institute of Technology Madras, Chennai, India \and
L3Cube Labs, Pune, India}

\maketitle

\begin{abstract}
Code-mixing is the practice of using two or more languages in a single sentence, which often occurs in multilingual communities such as India where people commonly speak multiple languages. Classic NLP tools, trained on monolingual data, face challenges when dealing with code-mixed data. Extracting meaningful information from sentences containing multiple languages becomes difficult, particularly in tasks like hate speech detection, due to linguistic variation, cultural nuances, and data sparsity. To address this, we aim to analyze the significance of code-mixed embeddings and evaluate the performance of BERT and HingBERT models (trained on a Hindi-English corpus) in hate speech detection. Our study demonstrates that HingBERT models, benefiting from training on the extensive Hindi-English dataset L3Cube-HingCorpus, outperform BERT models when tested on hate speech text datasets. We also found that code-mixed Hing-FastText performs better than standard English FastText and vanilla BERT models.

\end{abstract}
\section{Introduction}
Communication in India often involves using multiple languages, leading to code-mixing. This is the interchangeable use of vocabulary from two or more languages within sentences. Word embedding represents words in a continuous vector space, aiming to capture semantic relationships, syntactic patterns, and context information. Traditional word embeddings fail to capture the nuances of code-mixed languages, but newer models like FastText and BERT have overcome these shortcomings by utilizing subword information and analyzing entire sentences bidirectionally, respectively. These models capture the rich semantic meaning of the text.

The primary motivations for this paper include the need to address challenges posed by code-mixed text in NLP and the importance of code-mixed word embeddings by utilizing the aforementioned capabilities of FastText and BERT models. Another motivation is the need to demonstrate the advantages of BERT and Hing-BERT as well as FastText and Hing-FastText models, simultaneously providing a comparative analysis. The paper is organized into sections explaining the word embedding models used, literature survey, methodology, results and discussion, followed by the conclusion. 

\section{Word Embedding Models}
\subsection{BERT-Base}
BERT stands for Bidirectional Encoder Representations from Transformers, developed by Google AI Language in 2018. BERT revolutionized the NLP space by providing a solution for more than 11 main tasks in NLP, better than the traditional models. It is a bidirectional model that processes entire sentences at once, also providing a contextual understanding of it. BERT, with its 345 million parameters, was one of the first big NLP pre-trained models [5]. We have used BERT-base model on the datasets, which is pre-trained on huge amounts of English language data using a Masked Language Modeling (MLM) objective. 

\subsection{Multilingual BERT (mBERT)}
The Multilingual BERT model is used for this analysis, also being comparable to the HingBERT models. This mBERT model is available in the official BERT repository and it supports 104 languages, including Hindi and English languages [7]. It was pre-trained on the largest Wikipedia in a self-supervised manner. It was pre-trained on raw text only, without humans labeling them in any way.

\subsection{FastText Embedding Models}
FastText is an open-source library for text representation. It uses subword information to handle multilingual data and represents each word as a collection of letter n-grams, allowing it to capture semantic similarity and handle out-of-vocabulary words effectively. This approach enables FastText to generalize across languages and identify similarities and differences among words with shared letter n-grams, leading to improved performance in natural language processing tasks.

\subsection{L3Cube-HingCorpus and Hing Models}

\indent HingCorpus, developed by L3Cube in Pune, is a Hindi-English corpus containing 52.93 million sentences and 1.04 billion tags. This unsupervised HingCorpus is used to train Hing models such as HingBERT, Hing-mBERT, and Hing-RoBERTa, which provides a comparative analysis of BERT and HingBERT models. This Hing model is pre-compiled in HingCorpus, a mixture of Roman and mixed codes and types, but for this analysis we focus on Roman type.

We used several Hing models including HingGPT, HingFT and HingLID, for the datasets. HingGPT is a GPT2 transformer model trained on the extensive HingCorpus dataset using language modeling tasks. It's available in both Roman and Devanagari versions, expanding its range of applications. HingFT is a Hindi-English code-mixed FastText embedding model trained on Roman and Devanagari text from Hing-Corpus. HingLID is a code-mixed language identification model trained on a large in-house LID dataset.

\subsection{Hing Models vs. BERT models}

HingBERT models are designed to handle code-mixed datasets, effectively dealing with the unique characteristics for the same. They are pre-trained on real-world Hinglish text from Twitter, allowing them to better understand the intricacies of both languages. Compared to standard BERT models, HingBERT models are optimized to handle the specific challenges of Hinglish code-mixing and generate word embeddings that capture the meaning of words in both Hindi and English contexts.

\subsection{Hing FastText vs. vanilla FastText}
The Hing-FastText model, developed by L3Cube Pune, is trained on the L3Cube-HingCorpus dataset containing Roman and Devanagari text. Unlike traditional FastText models, it is designed for code-mixed Hindi-English data. It excels in tasks such as text classification, sentiment analysis, and hate speech detection due to its high context understanding of Hinglish text and its ability to handle diverse vocabularies that include words from both Hindi and English.

\section{Literature Survey}

In [1] the paper addresses hate speech detection in code-mixed text from Twitter, focusing on using transformer-based approaches like multilingual BERT and Indic-BERT. It explores single-encoder and dual-encoder settings, uses contextual text from parent tweets. Achieving an F1 score of 73.07 percent in the HASOC 2021 database demonstrates the importance of context and static embedding in hate speech classification. It also demonstrates the effectiveness of average representation in context-based text classification.

The [3] paper also explores hate speech detection in code-mixed text, comparing pre-trained embeddings like BERT, XLNet, and DistilBERT along with a CNN model for Hinglish code-mixed data. Despite BERT's popularity, XLNet has emerged as a strong performer in identifying hate speech. The study emphasizes the importance of using pre-trained embeddings for effective detection of hate speech in code-mixed environments.

The [10] introduces a model for hate speech detection in Hinglish, prevalent in social media conversations. It efficiently classifies tweets into hate-inducing, abusive, or non-offensive categories using character-level embeddings and GRU with Attention. The study highlights the need for robust moderation tools, especially in regions with diverse linguistic landscapes like India.

[2] explores hate speech detection in code-mixed Hinglish tweets on Twitter using transformer-based models like Indic-BERT, XLM-RoBERTa, and Multilingual BERT in a hard voting ensemble. It emphasizes the importance of considering conversation context in identifying hate speech and addresses limitations of mono-lingual classifiers. The paper achieves a macro F1 score of 0.7253 and suggests avenues for improvement, such as incorporating emojis and exploring better context understanding architectures.

\section{Methodology}

\subsection{Datasets}

In this study, the Hate Speech and Offensive Content (HASOC) dataset has been used as a basis to compare and evaluate the BERT and HingBERT models. There have been various versions of this dataset since 2019, containing languages like Hindi, English, Marathi, and German but we have leveraged the complexities of the Hindi-English dataset introduced in 2021. The HASOC Datasets have mostly been used in hate speech detection tasks in NLP. 

The Hate dataset includes a comprehensive collection of offensive text accumulated from Twitter. The dataset is divided into training, testing, and validation segments, all containing code-mixed text involving Hindi and English. The sentences contain various magnitudes of hate speech text, ranging from subjects encompassing politics, laws, crimes, popular culture, and so on.

\begin{table}[h]
\centering
\caption{Hate Speech Categories and Counts in a portion of Hate Dataset}
\renewcommand{\arraystretch}{1.1} 
\begin{tabular}{|p{0.45\textwidth}|c|c|}
\hline
\textbf{Hate Speech Category} & \textbf{Number of Examples}\\
\hline
Misogyny and Sexism & 249\\
\hline
Communal/Religious Hatred & 188\\
\hline
Incitement to Violence & 135\\
\hline
Victim-Blaming & 116 \\
\hline
\end{tabular}
\end{table}

\begin{table}[h]
\centering
\caption{Hate Speech Categories and Counts in a portion of Hasoc Dataset}
\renewcommand{\arraystretch}{1.1} 
\begin{tabular}{|p{0.45\textwidth}|c|c|}
\hline
\textbf{Hate Speech Category} & \textbf{Number of Examples}\\
\hline
Racial/Ethnic Hate & 248 \\
\hline
Gender-based Hate & 67 \\
\hline
Political Hate & 159 \\
\hline
Other Hate & 173 \\
\hline
\end{tabular}
\end{table}

\subsection{Preprocessing}
\subsection{Word Embedding Generation}

Our research methodology focuses on word embedding and preprocessing in NLP. We work with the datasets, preparing the data for training, testing, and validation. Word embedding creates dense vector representations of words, capturing their context. It's important for solving NLP problems and introduces vocabulary to learn word representations. Word embeddings help transform high-dimensional word spaces into low-dimensional vector spaces while preserving valuable information.


\begin{figure}[ht]
    \centering
    \includegraphics[scale=0.25]{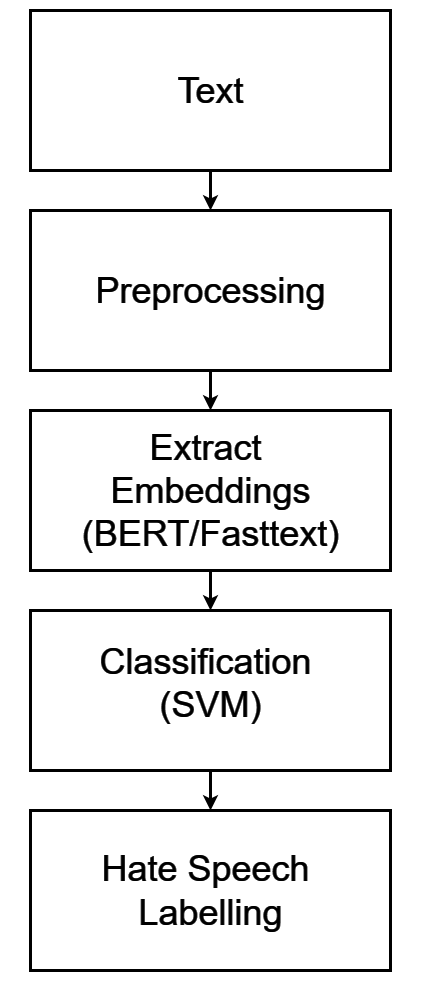}
    \caption{Flowchart}
    \label{fig}
\end{figure}

We are using a transformer model that can handle two languages, which brings a significant increase in efficiency and performance compared to previous methods. The models we are using include BERT, Multilingual BERT, DistilBERT, RoBERTa, HingBERT, Hing-mBERT, Hing-RoBERTa, HingGPT, HingBERT-LID and HingFT. The process involves encoding the input text sequence using the hidden state of the transformer architecture and applying an internal compression method to the number of inputs.

\subsection{Classification Algorithms}

After feature extraction, the classification model evaluates the word accuracy using various algorithms such as Random Forest, Linear Regression, SVM and KNN. The data is divided into training and test sets, and SVM is particularly useful for sentiment analysis due to its simplicity and efficiency with high-dimensional data.

The objective is to identify the best hyperplane that separates the training data using terms generated by BERT and HingBERT. Performance is evaluated on the training and evaluation sets, and metrics include F1-score, precision, accuracy, and recall.

\section{Results}
After assessing various BERT and HingBERT models on code-mixed datasets and obtaining results for the evaluation and testing sets, the testing scores are shown in Figures 2 and 3.
 We also studied pools and layers in the transformer model, mainly focusing on the last layer to obtain high-level input data. For this experimentation, mostly max pooling is used. Our research shows that the HingBERT model outperforms the BERT model, and likewise, Hing-FastText outperforms the vanilla FastText model with higher F1 scores, recall, and accuracy in identifying atypical words in scrambled text.


\begin{thebibliography}{00}
\bibitem{b1} Nayak, Ravindra and Raviraj Joshi. “Contextual Hate Speech Detection in Code Mixed Text using Transformer Based Approaches.” Fire (2021).

\bibitem{b2} Farooqi, Zaki Mustafa et al. “Leveraging Transformers for Hate Speech Detection in Conversational Code-Mixed Tweets.” Fire (2021).

\bibitem{b3}  I. Chaitanya, I. Madapakula, S. K. Gupta and S. Thara, "Word Level Language Identification in Code-Mixed Data using Word Embedding Methods for Indian Languages," 2018 International Conference on Advances in Computing, Communications and Informatics (ICACCI), Bangalore, India, 2018, pp. 1137-1141, doi: 10.1109/ICACCI.2018.8554501.

\bibitem{b4} L. Sravani, A. S. Reddy and S. Thara, "A Comparison Study of Word Embedding for Detecting Named Entities of Code-Mixed Data in Indian Language," 2018 International Conference on Advances in Computing, Communications and Informatics (ICACCI), Bangalore, India, 2018, pp. 2375-2381, doi: 10.1109/ICACCI.2018.8554918.

\bibitem{b5} S. Banerjee, B. Raja Chakravarthi and J. P. McCrae, "Comparison of Pretrained Embeddings to Identify Hate Speech in Indian Code-Mixed Text," 2020 2nd International Conference on Advances in Computing, Communication Control and Networking (ICACCCN), Greater Noida, India, 2020, pp. 21-25, doi: 10.1109/ICACCCN51052.2020.9362731.

\bibitem{b6} Arra’Di Nur Rizal and Sara Stymne. 2020. Evaluating Word Embeddings for Indonesian–English Code-Mixed Text Based on Synthetic Data. In Proceedings of the The 4th Workshop on Computational Approaches to Code Switching, pages 26–35, Marseille, France. European Language Resources Association.

\bibitem{b7} C. Sabty, M. Elmahdy, and S. Abdennadher, "Named Entity Recognition on Arabic-English Code-Mixed Data," in 13th International Conference on Semantic Computing (ICSC), Newport Beach, CA, USA, 2019, pp. 93--97.

\bibitem{b8} K. Reji Rahmath, P. C. R. Raj and P. C. Rafeeque, "Pre-trained Word Embeddings for Malayalam Language: A Review," 2021 International Conference on Artificial Intelligence and Smart Systems (ICAIS), Coimbatore, India, 2021, pp. 568-572, doi: 10.1109/ICAIS50930.2021.9396042.

\bibitem{b9} Adithya Pratapa, Monojit Choudhury, and Sunayana Sitaram. 2018. Word Embeddings for Code-Mixed Language Processing. In Proceedings of the 2018 Conference on Empirical Methods in Natural Language Processing, pages 3067–3072, Brussels, Belgium. Association for Computational Linguistics.

\bibitem{b10} C. Sabty, M. Elmahdy and S. Abdennadher, "Named Entity Recognition on Arabic-English Code-Mixed Data," 2019 IEEE 13th International Conference on Semantic Computing (ICSC), Newport Beach, CA, USA, 2019, pp. 93-97, doi: 10.1109/ICOSC.2019.8665500.

\bibitem{b11} N. Sabri, A. Edalat and B. Bahrak, "Sentiment Analysis of Persian-English Code-mixed Texts," 2021 26th International Computer Conference, Computer Society of Iran (CSICC), Tehran, Iran, 2021, pp. 1-4, doi: 10.1109/CSICC52343.2021.9420605.

\bibitem{b12} Rahul, V. Gupta, V. Sehra and Y. R. Vardhan, "Hindi-English Code Mixed Hate Speech Detection using Character Level Embeddings," 2021 5th International Conference on Computing Methodologies and Communication (ICCMC), Erode, India, 2021, pp. 1112-1118, doi: 10.1109/ICCMC51019.2021.9418261.

\bibitem{b13} Genta Indra Winata, Zhaojiang Lin, and Pascale Fung. 2019. Learning Multilingual Meta-Embeddings for Code-Switching Named Entity Recognition. In Proceedings of the 4th Workshop on Representation Learning for NLP (RepL4NLP-2019), pages 181–186, Florence, Italy. Association for Computational Linguistics.

\bibitem{b14} Pranaydeep Singh and Els Lefever. 2020. Sentiment Analysis for Hinglish Code-mixed Tweets by means of Cross-lingual Word Embeddings. In Proceedings of the The 4th Workshop on Computational Approaches to Code Switching, pages 45–51, Marseille, France. European Language Resources Association.






\end{thebibliography}
\begin{figure}[ht]
    \centering
    \includegraphics[scale=0.39]{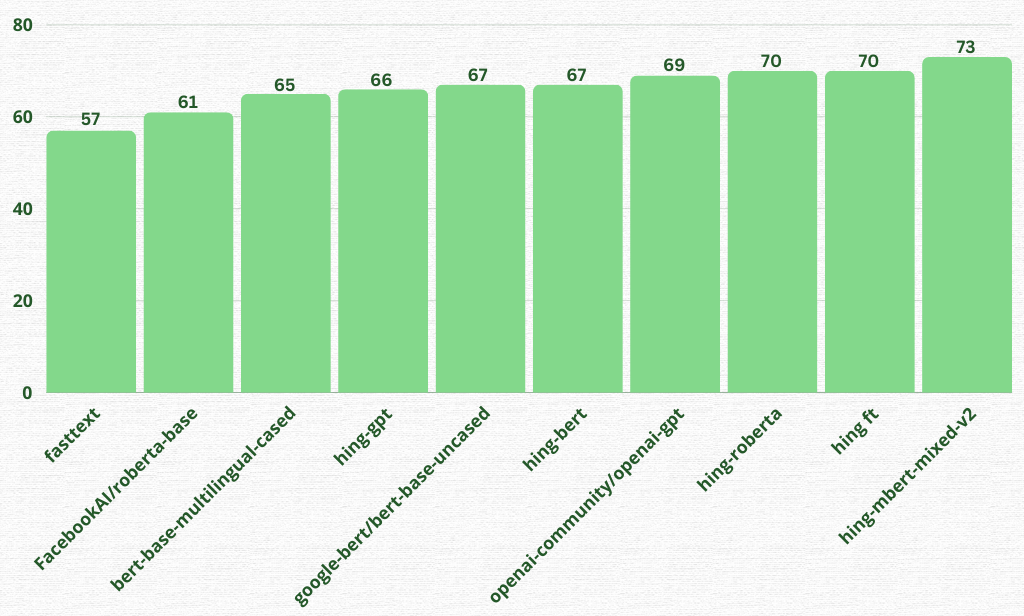}
    \caption{Comparison Chart for HASOC dataset}
    \label{fig:hasoc}
\end{figure}

\begin{figure}[ht]
    \centering
    \includegraphics[scale=0.39]{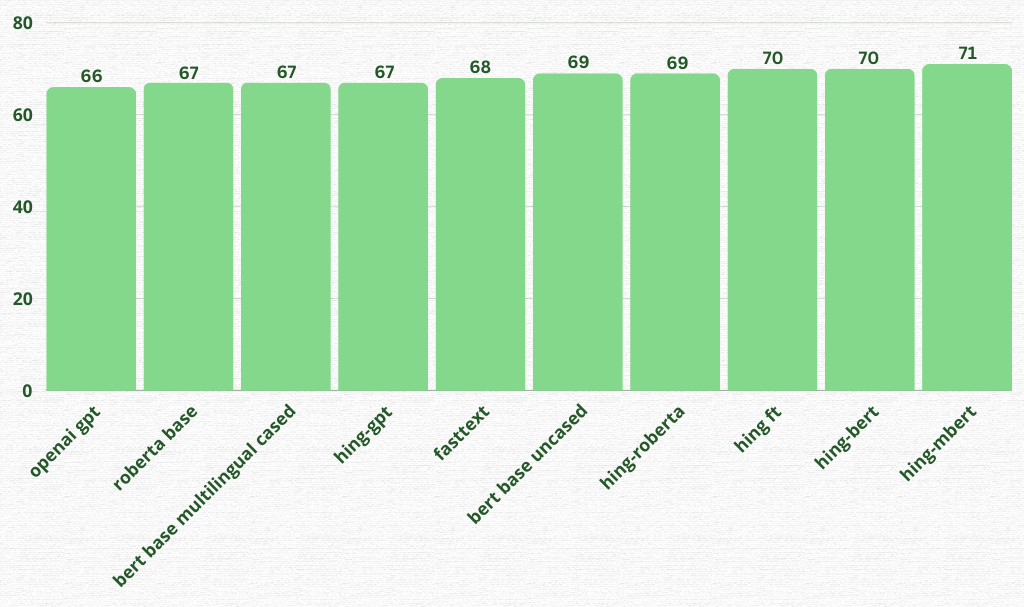}
    \caption{Comparison Chart for HATE dataset}
    \label{fig:hate}
\end{figure}

\section{Conclusion}

This paper discusses the use of specialized software to detect hate speech in multilingual contexts with a particular focus on Hindi-English (Hinglish) texts prevalent in India. This study demonstrates the superior performance of HingBERT trained on the L3Cube-HingCorpus database in detecting hate speech due to its ability to handle mixed code context. Additionally, the study emphasizes the need for specialized NLP models to address the challenges of data encoded in multilingual corpora.

We can expand the scope of this paper by using more diverse datasets, as the primary datasets may not capture the full range of linguistic variations in code-mixed Hinglish text. Also, the specialized Transformer and FastText models possess high computational complexity, which may be bettered by optimizing them. Further scope includes conducting real-world tests, to assess the usability of models used in a natural and practical scenario, like social media platforms. Exploring advanced NLP techniques like zero-shot learning, transfer learning may also prove in giving us better results.




\end{document}